\documentclass{article}

\usepackage{microtype}
\usepackage{graphicx}
\usepackage{subfigure}
\usepackage{booktabs} % for professional tables

\usepackage{hyperref}

\usepackage[accepted]{icml2023}

\usepackage{amsmath}
\usepackage{amssymb}
\usepackage{mathtools}
\usepackage{amsthm}
\newtheorem{defn}{Definition}
\newtheorem{thm}{Theorem}
\usepackage[capitalize,noabbrev]{cleveref}

\theoremstyle{plain}

\theoremstyle{definition}

\theoremstyle{remark}

\usepackage[textsize=tiny]{todonotes}

\icmltitlerunning{Analyzing Bias in Text-to-Image Models}

\begin{document}
\twocolumn[
\icmltitle{Word-Level Explanations for Analyzing Bias in Text-to-Image Models}



\icmlsetsymbol{equal}{*}

\begin{icmlauthorlist}
\icmlauthor{Alexander Lin}{equal,xxx}
\icmlauthor{Lucas Monteiro Paes}{equal,xxx}
\icmlauthor{Sree Harsha Tanneru}{equal,xxx}
\icmlauthor{Suraj Srinivas}{xxx}
\icmlauthor{Himabindu Lakkaraju}{xxx}
\end{icmlauthorlist}

% \icmlaffiliation{xxx}{Department of Computer Science, Harvard University, Cambridge, USA}
% \icmlaffiliation{yyy}{Department of Applied Mathematics, Harvard University, Cambridge, USA}
% \icmlaffiliation{zzz}{Institute for Applied Computational Science, Harvard University, Cambridge, USA}
\icmlaffiliation{xxx}{Harvard University, Cambridge, USA}

\icmlcorrespondingauthor{Sree Harsha Tanneru}{sreeharshatanneru@g.harvard.edu}
\icmlcorrespondingauthor{Lucas Monteiro Paes}{lucaspaes@g.harvard.edu}
\icmlcorrespondingauthor{Alexander Lin}{alexanderlin01@g.harvard.edu}

\icmlkeywords{Text-to-Image Generative Models, Explainability, Fairness}

\vskip 0.3in
]


% The  \icmlEqualContribution command is standard text for equal contribution.

\printAffiliationsAndNotice{\icmlEqualContribution -- authors in alphabetical order.} % otherwise use the standard text.

\begin{abstract}
%There is great interest in employing text-to-image models in many real world settings. 
Text-to-image models take a sentence (i.e. prompt) and generate images associated with this input prompt. These models have created award wining-art, videos, and even synthetic datasets.
%However, several works have reported that these models exhibit biases when they generate images. 
However, text-to-image (T2I) models can generate images that underrepresent minorities based on race and sex.
This paper investigates which word in the input prompt is responsible for bias in generated images.
We introduce a method for computing scores for each word in the prompt; these scores represent its influence on biases in the model's output.
Our method follows the principle of \emph{explaining by removing}, leveraging masked language models to calculate the influence scores. 
We perform experiments on Stable Diffusion to demonstrate that our method identifies the replication of societal stereotypes in generated images. 
%By gaining a deeper understanding of these biases, we pave the way for addressing the challenges in deploying text-to-image generative models equitably.
\end{abstract}

\section{Introduction}
\label{introduction}

Text-to-Image (T2I) models such as DALL-E \cite{DALLE}, Midjourney \cite{Midjourney}, and Stable Diffusion \cite{rombach2022high} have grown in popularity, and have been recently used to create award-winning art \cite{MoMa}, synthetic radiology images \cite{Chambon2022AdaptingPV}, and high-quality videos \cite{Xing21}. Broadly, T2I models take a text \emph{prompt} in natural language -- e.g., a sentence -- as input and generate an image associated with that prompt \cite{paiss2022no}.
%-- in this work we refer to the words in the prompt as \emph{words}. 

Recently, there have been growing concerns about the underrepresentation of minority groups in the images generated from T2I models.  In a recent Wired article \cite{wired}, when questioned about the launch of DALL-E 2, an external member of OpenAI ``Red Team" described their experience using the model as ``enough risks were found that maybe it shouldn't generate people or anything photorealistic." 

Underrepresentation of minority groups in T2I models has been rigorously analyzed \cite{cho2022dall, Xing21}. For example, \cite{cho2022dall} showed that the word ``likable'' was associated with lighter skin tones, while ``poor'' was associated with darker skin tones. Such biases are undesirable in these models because they lead to the underrepresentation of minorities which perpetuates discrimination \cite{An19}. While prior works identify biases in text-to-image models \cite{bianchi2022easily, Xing21, cho2022dall}, the causes of these biases in terms of problematic associations with words in input prompt has not been studied in previous works. Our work precisely aims to fill this gap by attributing bias in generated images to specific words in the input prompt.
% While \cite{bianchi2022easily, Xing21, cho2022dall} identify social biases in text-to-image models, and \cite{paiss2022no} made pixel-wise feature attribution in T2I models, our work systematically identifies input words that lead these models to propagate underrepresentation in the produced images.
%no work has been done on  systematically identifying input words that cause T2I models to generate biased images.
% This bias leads to underrepresentation of minorities, leading to the propagation of historical discrimination in our society [ref]. 

The underrepresentation in the model's output motivates our main question: 
    \textbf{``Which word in the prompt caused underrepresentation?''}
For example, consider the prompt ``A respected doctor at the hospital.'' 
In Fig. \ref{fig::Diagram}, we observe that Stable Diffusion v.1.4 mainly generates images of male doctors for this prompt. 
For each word in the prompt, our method calculates how responsible that word is for the observed bias. 
We show that the word \emph{doctor} is responsible for the underrepresentation of females in the model's output. 
Answering such questions allows practitioners to (i) identify the root of the bias in their models and (ii) modify prompts to achieve better output representation.

The \textbf{main contributions} of this work are:
    \textbf{(1)} We propose a word-influence metric that encodes the influence of a given word in the underrepresentation of the model's output. Our metric can be used to determine which word in the prompt causes underrepresentation and guide practitioners on devising measures to alleviate bias.
    \textbf{(2)} We propose a model-agnostic method to evaluate our metric for a prompt. Our word-influence method is inspired by leave-one-out while maintaining the semantic coherence of prompts during the word-influence evaluation process.   
    \textbf{(3)} We run experiments on Stable Diffusion v.1.4 and show that our metric captures the bias associated with words in a prompt.
    %and its interactions in a given prompt.

% The code for our experiments can be found \href{https://github.com/harsha070/CS282BR}{here}.

\subsection{Related Work}
\label{relatedwork}

\textbf{Feature Importance in Text Classification.} As machine learning models become more complex, the ability to explain its predictions is required to engender user trust and provide insights for model improvement. 
LIME \cite{ribeiro2016should} interprets individual predictions based on locally approximating a model. 
SHAP \cite{lundberg2017unified}, based on cooperative game theory, assigns each feature an importance value for a particular prediction. 
\cite{covert2022explaining} proposes a framework based on the idea of simulating feature removal to quantify each feature's influence. 
These method explain a classification model's predictions, however, they are not tailored to explain generative models. 
Our work fills this gap by providing a method, inspired by SHAP, to analyze bias and word influence attribution in generative models.
%they do not specifically analyze bias in sensitive attributes in generative models.

\textbf{Fairness and Explainability in Text to Image Models.} Studies \cite{bianchi2022easily, Xing21, cho2022dall, paiss2022no} have raised concerns about how generative models perpetuate and amplify social biases. \citet{bianchi2022easily} 
%investigates image generation models that are easily available online, and 
found that certain prompts perpetuate stereotypes based on sensitive attributes and amplify social disparities. 
\citet{cho2022dall} suggested that the skin color and sex of images generated by T2I models heavily depend on words agnostic to these sensitive attributes.
\citet{paiss2022no} studies the relation between the prompt and the image by explaining how individual pixels are related to words within a text prompt.
Our work differs from the previous ones by analyzing which word in the input prompt was responsible to generated the biases found in \cite{bianchi2022easily, Xing21, cho2022dall}.

\section{Word-Influence for Representativeness}
\label{sec:method}

\begin{figure*}[t]
    \centering
    \includegraphics[width=.7\linewidth]{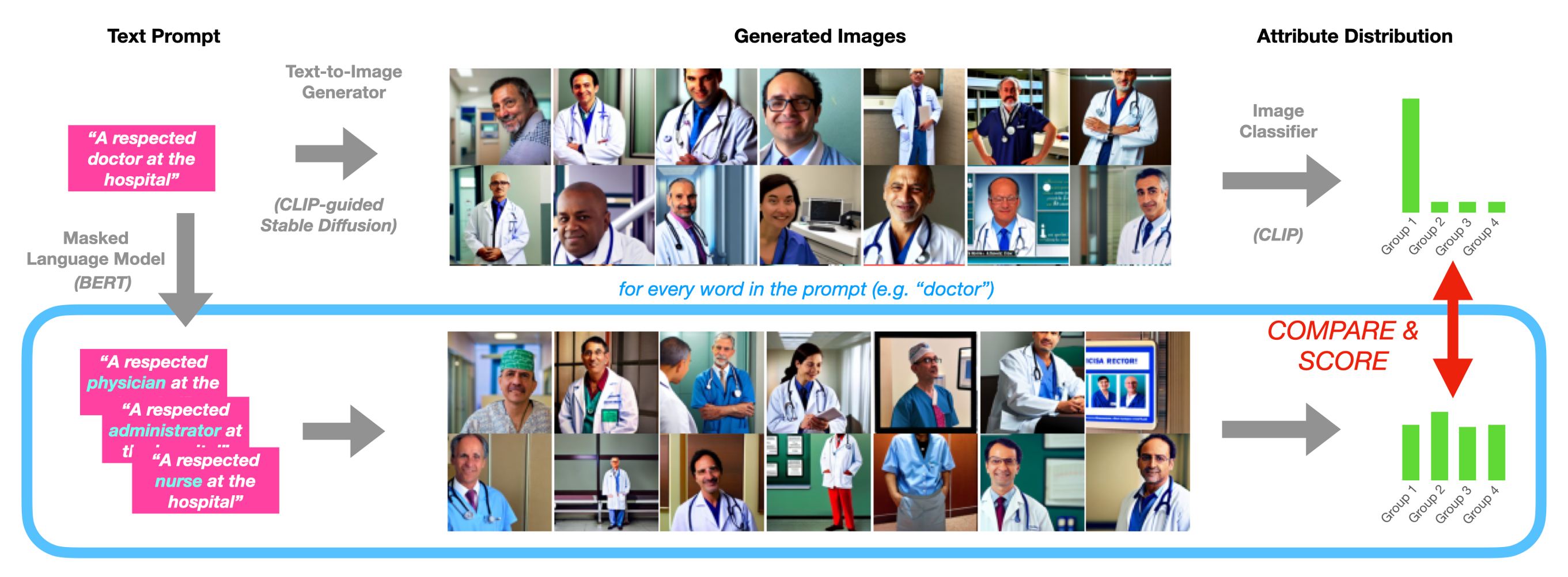}
    \caption{The diagram showing the proposed pipeline for the input prompt \emph{``A respected doctor at the hospital''}.
    }
    \label{fig::Diagram}
\end{figure*}

\textbf{Problem Setup and Notation.}
Let $p = p_1, p_2, \ldots, p_k \in \mathcal{P}$ denote a text prompt (\emph{sentence}) comprised of words $p_1, p_2, \ldots, p_k$ where $k$ is the number of words in the prompt. A text-to-image generative model is denoted by $\textsc{T2I}$, i.e., $\textsc{T2I}(p)$ is a generated sample from a distribution of images $\bold x \in \mathcal{X}$ as a function of the prompt $p$. Moreover, let a sensitive attribute (e.g., sex, race, and age) of a person in an image be given by $\mathcal{G}(\bold x) \in \{\bold g_1, ..., \bold g_l\} \triangleq \mathbb{G}$ -- e.g., the sex of a person in the image $\bold x$. Finally, $\Pr_{T2I(p)}[\mathcal{G}(X) = \bold g]$ denotes the probability of the sensitive attribute of the generated images to be $\bold g$. 
%We will define a word influence score $TI(p, i, \bold g)$ for each word $p_i$ in a prompt $p$ for a sensitive attribute $\bold g$.

\textbf{Word impact on representativeness.} We start by considering the original prompt \emph{``A respected doctor at the hospital"} showed in Fig. \ref{fig::Diagram}.
Using CLIP-guided Stable Diffusion, this prompt generates images that are mostly males.
%We make this observation by analyzing the group distribution of the generated images on the right-hand side of Fig. \ref{fig::Diagram}.
However, the biased group distribution may be a consequence of the words ``respected", ``doctor," or even `` hospital," next, we analyze which word is responsible for it.

To understand the impact of each word in the group distribution of the generated images, ideally, we would remove words from the prompt and analyze how this changes the sex of the produced images -- as in explaining by removing \cite{covert2022explaining}. 
However, word removal may lead to sentences that are not grammatically coherent. 
For example, by removing the word ``doctor," the original prompt becomes ``A respected at the hospital" -- this sentence has no grammatical coherence.
Therefore, we will replace each word with multiple other candidates generated by BERT \cite{devlin2018bert}.
BERT ensures that the sentence with replaced word has a rough grammatical coherence.

In Fig. \ref{fig::Diagram}, we show the images generated by the T2I model when we replace the word \emph{``doctor"} in the original prompt with other words such as ``physician," ``administrator," and ``nurse." 
By replacing \emph{``doctor"} the sex of the model generates images has more representativiness.  
%with more group representativeness by making these word changes. 
Therefore, we ask the question \textbf{How does the group distribution change when we replace certain words in the original prompt?}
To answer this question, we compare the distribution of the original prompt with the distribution of groups generated by the modified prompt. 
To make this comparison, it is necessary to have a systematical method to generate group distributions from a given (and transmuted) prompts.  
Next, we propose a method to do so. 

\textbf{Proposed Pipeline.} We define a pipeline to (i) generate coherent transmuted prompts, (ii) sample images using the prompts, and (iii) attribute each image to a sensitive group. Our pipeline has three components, illustrated in Fig. \ref{fig::Diagram}.
\begin{enumerate}
    \item \textbf{Text Transmuter:}
    We use a pretrained masked language model (MLM) to replace words in the original prompt. The MLM will substitute a word that is different from the original one, but still roughly obeys grammatical rules and completes the text in a sensical manner. In our implementation, we use a BERT-base MLM \citep{devlin2018bert}.     
    \item \textbf{Image Generator:} Next, we pass each of the candidate prompts (the prompt with a replaced word) through the image generator and sample one image per prompt. This set of samples captures the distribution of images \emph{conditioned on removal the  $i$-th word}. Thus, it provides a counterfactual image distribution (i.e. what would have happened if word $i$ did not exist in the original prompt). Here we use CLIP-guided Stable Diffusion \citep{rombach2022high} for image generation. 
    
    \item \textbf{Group Classifier:}  
    Finally, we use a classifier that takes an image $\bold x$ as input and classifies it as  a member of a group $\bold g$.  
    In our pipeline, we use CLIP as a classifier by assigning $\bold x$ to the group that maximizes the CLIP score $\bold g = \arg \max_{\bold g' \in \mathbb{G}} \text{CLIP}(\bold x, \texttt{text}(\bold g'))$, where $\texttt{text}(\bold g')$ means to write $\bold g'$ as a text string (e.g. ``male'' or ``female'').\footnote{CLIP is also used by the image generator, therefore it may be biased when classifying images.
    %the image this way (i.e. we might be afraid that CLIP latches onto features that have nothing to do with gender in classifying the image). 
    %While this is a valid concern, 
    However, we found otherwise -- see appendix \ref{clip1} \ref{clip2} for the discussion.
    %through preliminary comparisons with manual classification that the predictions given by CLIP are actually quite accurate (see Tables \ref{clip1} and \ref{clip2} in the Appendix).
    } 
\end{enumerate}

% \begin{figure*}[t]
%     \centering
%     \includegraphics[width=.6\linewidth]{images/Diagram.png}
%     \caption{The diagram showing the proposed pipeline for \emph{``A respected doctor at the hospital''}.
%     }
%     \label{fig::Diagram}
% \end{figure*}

Our pipeline is model-agnostic and not restricted to the particular model choices we made here.
With our proposed pipeline, we have access to the group membership distribution of the generated images. 
Hence, we are able to measure the importance of each word in the representativeness -- next we define the word influence. 

\textbf{Word influence.}
% What do we aim to do?
We define a \emph{word-influence score} $TI^{\mathcal{G}}(p, i, \bold g): \mathcal{P} \times [k] \times \mathbb{G} \rightarrow \mathbb{R}$.
%a function such that, given a group membership of interest $g$ (e.g., sex = Male), can attribute a value for each word $p_i$ in the prompt $p \in \mathcal{P}$ that represents this word influence on $\mathcal{G}(\bold x) = \bold g$.
Intuitively, this word-influence score formalized in Definition \ref{def::TIS} quantifies how much each word $p_i$ is responsible for producing the property $\mathcal{G}(\bold x) = \bold g$ within the images generated by the prompt $p$. 
If $TI^{\mathcal{G}}(p, i, \bold g) > TI^{\mathcal{G}}(p, j, \bold g)$ the word $p_i$ influences more $\mathcal{G}(\bold x)$ than $p_j$ for the prompt $p$.
When $\mathcal{G}$ is clear from the context, we denote $TI^{\mathcal{G}}(p, i, \bold g)$ by $TI(p, i, \bold g)$.

We denote the probabilities of a generated image being part of a given group $\bold {g}$ by:
\begin{align*}
    P_{\mathcal{S}}(\bold g) & \triangleq \Pr_{\textsc{T2I}(\textsc{BERT}(p_{/\mathcal{S}}))}[\mathcal{G}(X) = \bold g]\\
    P_{\mathcal{S} \cup \{i\}}(\bold g) & \triangleq \Pr_{\textsc{T2I}(\textsc{BERT}(p_{/\mathcal{S} \cup \{i\}}))}[\mathcal{G}(X) = \bold g]
\end{align*}
where $\text{BERT}(p_{/\mathcal{S}})$ means to replace the words corresponding to subset $\mathcal{S}$ using BERT \cite{devlin2018bert}. Next, we define the word Influence Score.

\begin{defn}[word Influence for group $\bold g$] Given a prompt $p$ with $k \in \mathbb{N}^+$ words, we define the $r$-level influence of word $i \in [k]$ for the group $\bold g$ as: 
% \begin{equation} 
%     TI(p, i, r, \bold g) = \sum_{\substack{\mathcal{S} \subseteq [k] \\ i \not \in \mathcal{S} \\  |\mathcal{S}| \leq r-1}} \frac{\Pr_{\textsc{T2I}(\textsc{BERT}(p_{/\mathcal{S}}))}[\mathcal{G}(X) = \bold g] - \Pr_{\textsc{T2I}(\textsc{BERT}(p_{/\mathcal{S} \cup \{i\}}))}[\mathcal{G}(X) = \bold g]}{{k-1 \choose |\mathcal{S}|}},
% \end{equation}
\begin{equation}
     TI(p, i, r, \bold g) \triangleq \sum_{\substack{\mathcal{S} \subseteq [k] \\ i \not \in \mathcal{S} \\  |\mathcal{S}| \leq r-1}} \frac{P_{\mathcal{S}}(\bold g) - P_{\mathcal{S} \cup \{i\}}(\bold g)}{{k-1 \choose |\mathcal{S}|}}, 
\end{equation}
% \begin{equation} 
% \begin{split}    
%     TI(p, i, r, \bold g) = \sum_{\substack{\mathcal{S} \subseteq [k] \\ i \not \in \mathcal{S} \\  |\mathcal{S}| \leq r-1}} \frac{1}{{k-1 \choose |\mathcal{S}|}} \left\{ \Pr_{\textsc{T2I}(\textsc{BERT}(p_{/\mathcal{S}}))}[\mathcal{G}(X) = \bold g] - \\ \Pr_{\textsc{T2I}(\textsc{BERT}(p_{/\mathcal{S} \cup \{i\}}))}[\mathcal{G}(X) = \bold g] \right\},
% \end{split}
% \end{equation}

SHAP values inspired our definition for the word influence \cite{lundberg2017unified}.
\label{def::TIS}
\end{defn}

\textbf{Measuring Word influence.} At first, our definition of the word influence score may seem impractical because it is a function of the group probability distribution, which is unknown -- imagine knowing the race distribution of images generated by a large language model. However, the word influence score may be approximated by generating multiple images using the same prompt. 
We show in Theorem \ref{thm::wordInfluenceConcentrates} that our estimation for the word-influence converges exponentially fast to the true word-influence score. 

%Let $p \in \mathcal{P}$ be a prompt. 
For each distribution $P_{\mathcal{S}}$ and $P_{\mathcal{S} \cup \{i\}}$ we sample $m$ i.i.d. images 
%$\{X_1^{/\mathcal{S}}, X_2^{/\mathcal{S}}, ..., X_m^{/\mathcal{S}}\}$ and $\{X_1^{/\mathcal{S} \cup \{i\}}, X_2^{/\mathcal{S} \cup \{i\}}, ..., X_m^{/\mathcal{S} \cup \{i\}}\}$. 
and denote the empirical distribution of $\bold g$ in the images 
%or the probability of the image to be a member of a group $\bold g$ as 
$\hat{P}_{\mathcal{S}}(\bold g) $ and $\hat{P}_{\mathcal{S} \cup \{i\}}(\bold g)$ -- see appendix for details. With this, we define the approximation for the word-influence for group $\bold g$ as:

% \begin{equation}
%     \widehat{TI}(p, i, r, \bold g) = \sum_{\substack{\mathcal{S} \subseteq [k] \\ i \not \in \mathcal{S} \\  |\mathcal{S}| \leq r-1}} \frac{\widehat{ \Pr}_{\textsc{T2I}(\textsc{BERT}(p_{/\mathcal{S}}))}[\mathcal{G}(X) = g]- \widehat{ \Pr}_{\textsc{T2I}(\textsc{BERT}(p_{/\mathcal{S}  \cup \{i\}}))}[\mathcal{G}(X) = g]}{{k-1 \choose |\mathcal{S}|}}.
%     \label{eq::tiest}
% \end{equation}

\begin{equation}
    \widehat{TI}(p, i, r, \bold g) = \sum_{\substack{\mathcal{S} \subseteq [k] \\ i \not \in \mathcal{S} \\  |\mathcal{S}| \leq r-1}} \frac{\hat{P}_{\mathcal{S}}(\bold g) - \hat{P}_{\mathcal{S} \cup \{i\}}(\bold g)}{{k-1 \choose |\mathcal{S}|}}
    \label{eq::tiest}
\end{equation}

\begin{thm}[Shap Word Influence Concentrates]
Let $\widehat{TI}(p, i, r, \bold g)$ be the estimator for the word-influence defined in \eqref{eq::tiest}. 
If the sampled images are i.i.d. using the image generator for the prompts $P_{\mathcal{S}}$ and $P_{\mathcal{S} \cup \{i\}}$ then:
% \begin{equation}
% \resizebox{9cm}{!}{$
%     \Pr[| \widehat{TI}(p, i, r, \bold g) - {TI}(p, i, r, \bold g)| > t] \leq 4 \sum_{\substack{\mathcal{S} \subseteq [k] \\ i \not \in \mathcal{S} \\  |\mathcal{S}| \leq r-1}}  \exp^{-\frac{t^2 m {k-1 \choose |\mathcal{S}|}^2 }{ (\sum_{|\mathcal{S}|}  {k-1 \choose |\mathcal{S}|})^2 }}.$}
% \end{equation}
\begin{equation}
    \Pr[| \widehat{TI}(p, i, r, \bold g) - {TI}(p, i, r, \bold g)| > t] \leq O(e^{-m})
\end{equation}

    \label{thm::wordInfluenceConcentrates}
\end{thm}

Now that we have shown that it is possible to estimate the word-influence in Definition \ref{def::TIS} via resampling images, in the next section, we show empirical results using this metric.  

\section{Experimental Results}

% \begin{itemize}
%     \item "The ceo promp" --> 2 graphs
%     \item "the substitution prompt" --> word influence graph. 
%     \item Transmutations table
%     \item 
% \end{itemize}

\textbf{A Detailed First Example.}
\label{sec:first}
%To illustrate how our framework works, we begin with a detailed first example: 
Consider the prompt \emph{``a respected doctor at the hospital''}.  
% Figure \ref{fig::Diagram} shows sample images generated by CLIP-guided Stable Diffusion (version 1.4), one of the state-of-the-art text-to-image generative models.  
% We observe that the vast majority of generated images contain male individuals.  We now use our pipeline to understand which words in the original image contribute to this bias (e.g. ``respected'', ``doctor'', ``hospital'', or something else). 
Figure \ref{fig::Diagram} shows that by using Stable Diffusion, the majority of generated images contain male individuals.
Using our pipeline we can detect which words in the input prompt contribute to the underrepresentation of females given by Table \ref{word-influences}.
%(e.g. ``respected'', ``doctor'', ``hospital'', or something else).
%We test our framework on the prompt \emph{``a respected doctor at the hospital''}.  
% These are the experimental settings: We generate images of size $384 \times 384$ and use 30 iterations of Stable Diffusion.  For the original prompt, we generate 50 images.  We compute $r=1$-level word influence scores, which means we mask words one at a time.  With BERT, we generate 15 candidates per masked word, and therefore, a distribution of 15 images for each word that is removed.  Classifications are made using the CLIP classifier, as previously explained in Section \ref{sec:method}.  
%The final result of applying our pipeline to the aforementioned prompt is given 
%By applying our pipeline, we compute the quantities in Table \ref{word-influences}. 
From the word influence calculated for each word, we observe that the word ``doctor'' has the highest score for male, followed by ``respected". 
The word ``hospital" has slight female bias. 
The prepositions and articles (e.g. ``at'', ``the'', ``a'') have negligible scores because they do not affect the semantic meaning of the sentence -- see appendix \ref{table:trasmutations} for examples of text transmutations.
%To provide further insight into our method, we also provide Table \ref{table:trasmutations} (in the Appendix) to present some sample transmutations from masking words within the prompt.

% \begin{table}[h]
%     \centering
%     \resizebox{9cm}{!}{%
%     \begin{tabular}{|c|c|c|c|}
%     \hline
%     Prompt ($p$) & $P(\text{Male} | p)$ & $P(\text{Female} | p)$ & word Influence\\
%     \hline
%     Original Prompt	& 0.840 & 0.160 & -----\\
%     \hline
%     Prompt without ``a" & 0.867 & 0.133 & -0.027 \\
%     Prompt without ``respected" & 0.733 & 0.267 & +0.107\\
%     Prompt without ``doctor"	& 0.467 & 0.533 & \textbf{+0.373}\\
%     Prompt without ``at"	& 0.800 & 0.200 & +0.040\\
%     Prompt without ``the" & 0.800 & 0.200 & +0.040\\
%     Prompt without ``hospital" & 1.000 & 0.000 & -0.160\\
%     \hline
%     \end{tabular}
%     }
%     \vspace{1em}
%     \caption{The probability of (being as classified as) male vs. female for (a) the original prompt ``a respected doctor at the hospital''and  and (b) the original prompt with each word replaced.  For each word $i$, the word influence is calculated as the difference between $P(\text{Male} | p)$ and $P(\text{Male} | p_{/i})$.}
%     \label{word-influences}
% \end{table}

\begin{table}[t]
    \centering
    \resizebox{8cm}{!}{%
    \begin{tabular}{|c|c|c|c|}
    \hline
    Prompt ($p$) & $P(\text{Female} | p)$ & word Influence\\
    \hline
    Original Prompt	& 0.160 & -----\\
    \hline
    Prompt without ``a"  & 0.133 & -0.027 \\
    Prompt without ``respected" & 0.267 & +0.107\\
    Prompt without ``doctor"	 & 0.533 & \textbf{+0.373}\\
    Prompt without ``at"	 & 0.200 & +0.040\\
    Prompt without ``the"  & 0.200 & +0.040\\
    Prompt without ``hospital"  & 0.000 & -0.160\\
    \hline
    \end{tabular}
    }
    \vspace{1em}
    \caption{Word influence for each word and the probability of being classified as female for the original prompt ``a respected doctor at the hospital'' and replacing each word. 
    %For each word $i$, the word influence is calculated as in Definition \ref{def::TIS}.}
    %the difference between $P(\text{Male} | p)$ and $P(\text{Male} | p_{/i})$.
    }
    \label{word-influences}
\end{table}

\textbf{The Effect of Multiple Shapley Levels.}
We now provide a second example to demonstrate the utility of computing $r$-level word influence scores for $r > 1$.  As explained in Section \ref{sec:method}, this is inspired by a connection to the Shapley values framework \citep{lundberg2017unified}.
%, a well-established technique in explainable AI. 
We consider the following prompt: \emph{``the ceo of a fortune 500 company''}. 

\begin{figure}[b]
    \centering
    \includegraphics[width=0.49 \linewidth]{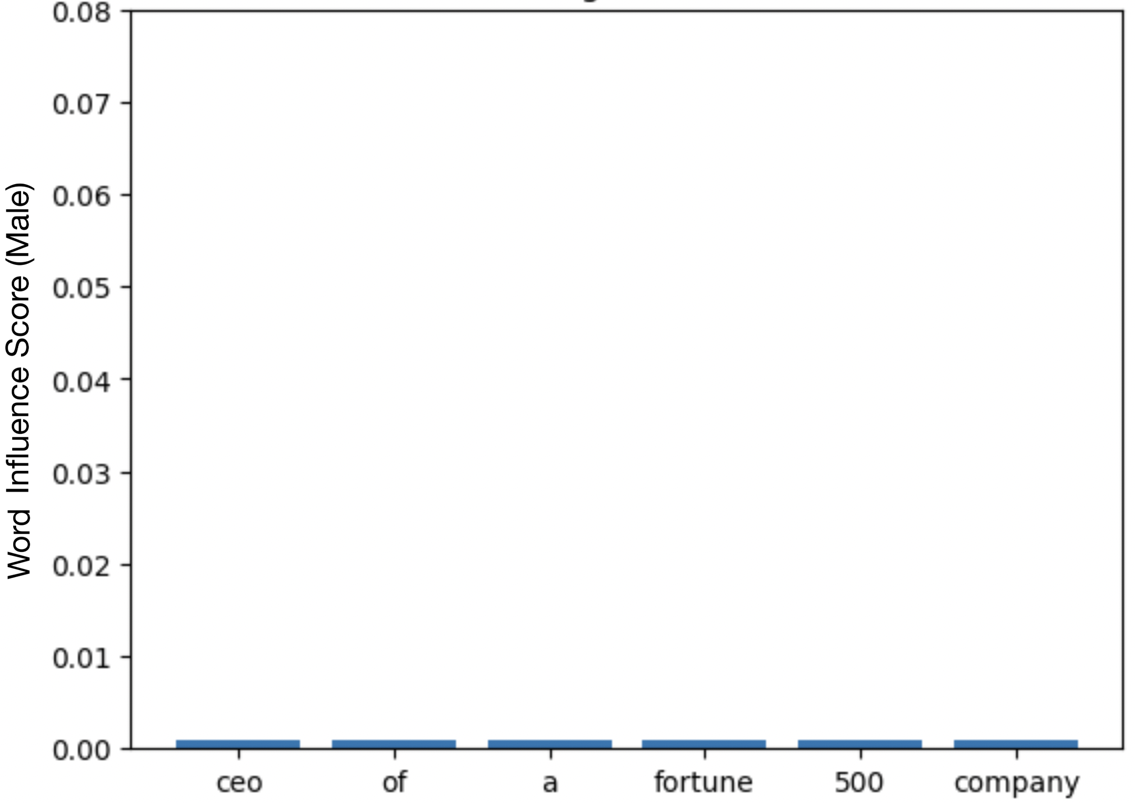}
    \includegraphics[width=0.49 \linewidth]{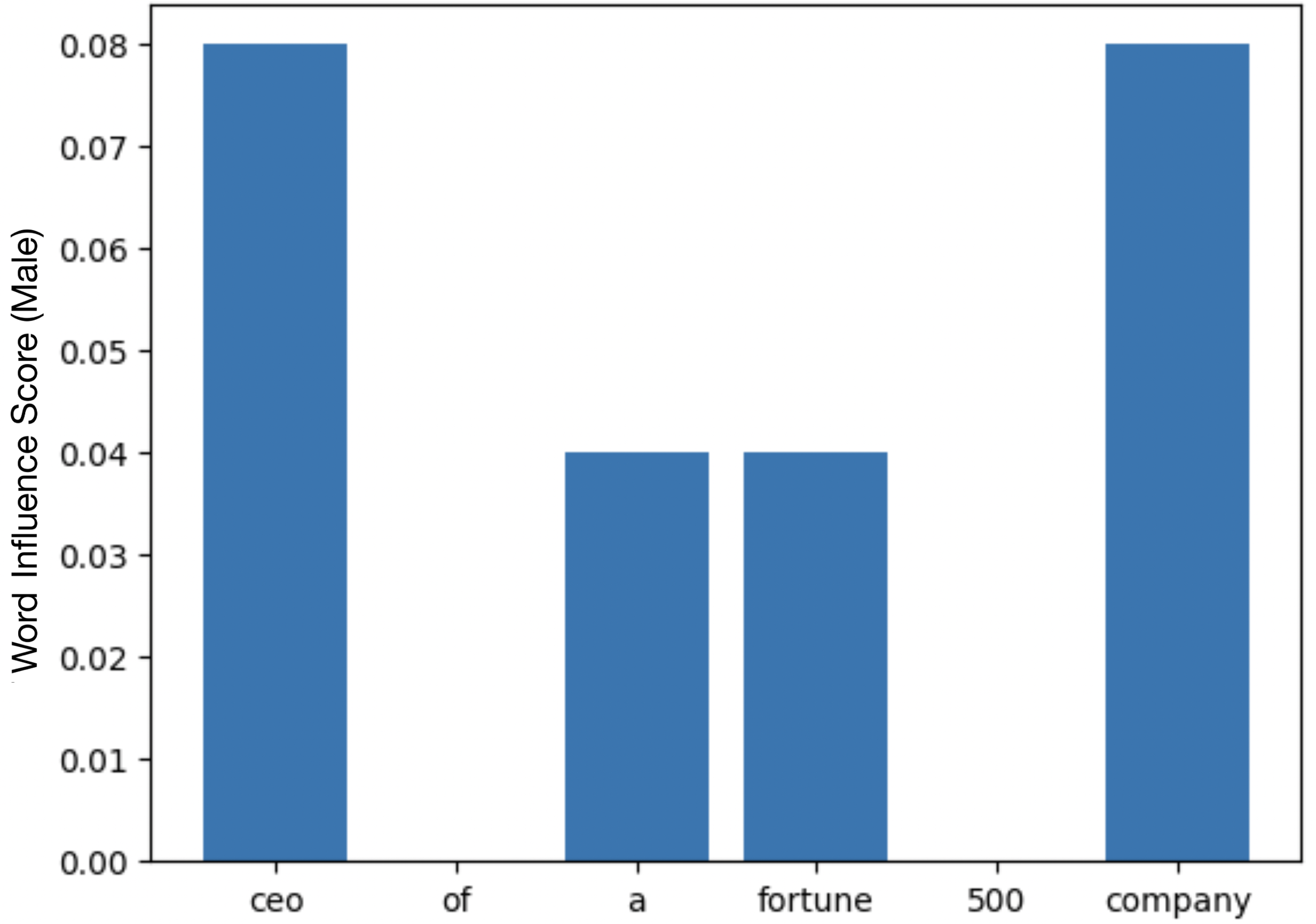}
    \caption{Word influence per word in the input prompt "ceo of a fortune 500 company" using $(r = 1)$ (left) and $(r = 2)$ (right) as in Definition \ref{def::TIS}.}
    \label{fig:shap}
\end{figure}

Figure \ref{fig:shap} (left) presents $r=1$-level word influence scores for this prompt, following the same setup as that of Section \ref{sec:first}.  We observe that all words have zero influence score.  This is because (i) the original prompt only generates male images and (ii) no matter which single word is removed from the original prompt, the remaining words still carry a heavy male bias, so the sex distribution does not change.  Thus, $r=1$ leads to uninformative scores for this prompt.        

In contrast, Fig. \ref{fig:shap} (right) presents $r=2$-level word influence scores for the same prompt, which considers replacement of all subsets of size $\leq 2$.  We observe that some words have non-zero scores. 
Notably, the words ``ceo'' and ``company'' have the largest effect on the sex of the images being male.  This indicates that replacing multiple words in the input prompt produces more informative scores by considering complex interactions between words.  
%because it enables better estimation of the marginal contribution of each word.  
However, increasing $r$ also leads to higher computational costs.
%(e.g. in our example prompt, going from $r = 1$ to $r = 2$ increases the number of subsets by three-fold, requiring three times the amount of computation).  

\begin{figure}[t]
    \centering    \includegraphics[width=\linewidth]{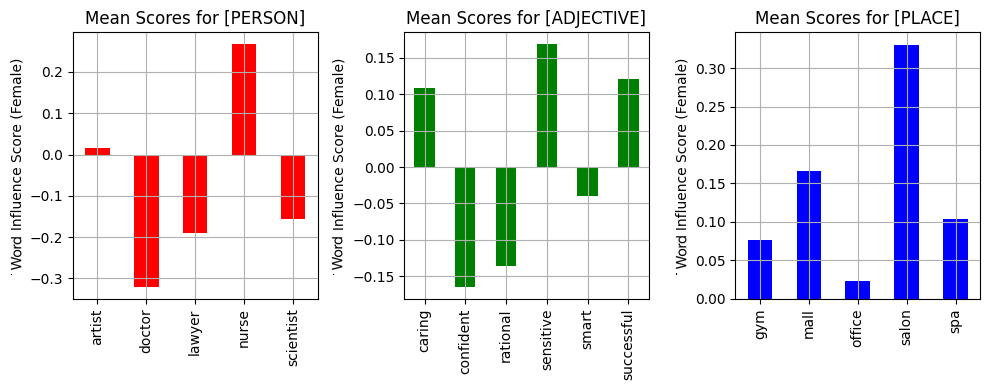}
    \caption{Averaged word influence scores for the words [PERSON] (\textbf{left}), [ADJECTIVE] (\textbf{center}), and [PLACE] (\textbf{right}) across all prompts in the large-scale evaluation.}
\label{fig:largescale}
\end{figure}

\textbf{Large-Scale Evaluation.} \label{sec:large-scale}
Finally, we test our pipeline on a collection of prompts to demonstrate how insights on model bias can be extracted.
We create 150 prompts with the following structure: \text{\emph{``a [ADJECTIVE] [PERSON] at the [PLACE]"}} (e.g., ``a confident doctor at the mall'')
% \begin{align*}
% \text{\emph{``a [ADJECTIVE] [PERSON] at the [PLACE]"}}
% \end{align*}
where values for $\text{\emph{[ADJECTIVE]}}$, $\text{\emph{[PERSON]}}$, and $\text{\emph{[PLACE]}}$ are given in the Appendix. For each prompt, our pipeline gives influence scores for every word.  %We use the same settings as Section \ref{sec:first}, but generate 10 images per mask to reduce computational costs.  
In Figure \ref{fig:largescale}, we show the average influence score across all prompts containing that word. We observe that words such as ``nurse'', ``caring'', ``sensitive'', and ``salon'' are associated with female, i.e., their inclusion in the prompt leads to the generation of female images.  
Similarly, the words ``doctor'', ``scientist'', ``confident'', and ``rational'' lead to the generation of male images.

\section{Final Discussion}

\textbf{Conclusion.} There are growing concerns that text-to-image (T2I) systems perpetuate and amplify stereotypes about minorities.
In this work, we provide a method to calculate the relative importance of each word for the representativeness of individuals with a given sensitive attribute. 
Specifically, given a prompt and a text-to-image model, our approach assigns a score to each word in the prompt, representing its impact on the number of images of individuals with a given sensitive attribute. 
% Our method can be used to design prompts that produce diverse images.
Moreover, our approach can be used to study if a T2I model associates sensitive attributes to words that are agnostic to them.
Our results indicate that Stable Diffusion associates words like ``scientist'' and ``lawyer'' with males while associating ``salon'' and ``sensitive'' with females.
%We noticed that our approach of computing score of each word by substitution, falls short in some prompts where there are interactions between words. We move to $r$-level SHAP score as a better metric to compute word influence scores and capture interactions between words. 
%We hope that this work motivates the inclusion of fairness considerations while building T2I systems.

\textbf{Limitations.}  
%While our method can be extended to analyse bias in any sensitive attribute, the results shown are restricted to sex alone. We leave the analysis of other protected attributes (e.g., race, ethnicity, etc) for future work.
While our proposed pipeline admits \emph{any} text transmutation algorithm, define the best way to generate them  is still an open question. 
In this paper, we use BERT to generate replacement candidates for each word. 
Alternative approaches are (i) the use of other masked language models, (ii) considering all possible syntactically correct word candidates, and (iii)  
an exhaustive list of synonyms and antonyms as word replacements.
%And finally, computing SHAP scores is computationally expensive, even more so, with the added image generation process.

\nocite{langley00}

\bibliography{main}
\bibliographystyle{icml2023}

%%%%%%%%%%%%%%%%%%%%%%%%%%%%%%%%%%%%%%%%%%%%%%%%%%%%%%%%%%%%%%%%%%%%%%%%%%%%%%%
%%%%%%%%%%%%%%%%%%%%%%%%%%%%%%%%%%%%%%%%%%%%%%%%%%%%%%%%%%%%%%%%%%%%%%%%%%%%%%%
% APPENDIX
%%%%%%%%%%%%%%%%%%%%%%%%%%%%%%%%%%%%%%%%%%%%%%%%%%%%%%%%%%%%%%%%%%%%%%%%%%%%%%%
%%%%%%%%%%%%%%%%%%%%%%%%%%%%%%%%%%%%%%%%%%%%%%%%%%%%%%%%%%%%%%%%%%%%%%%%%%%%%%%
\newpage
\appendix
\onecolumn
\section{Appendix}

\subsection*{Approximation of the Word Influence Score}
We define the empirical probabilities to approximate the word influence in section \ref{sec:method} as:
\begin{align}
   \widehat{ \Pr}_{\textsc{T2I}(\textsc{BERT}(p_{/\mathcal{S}}))}[\mathcal{G}(X) = g] & \triangleq \frac{1}{m} \sum_{j = 1}^{m} 1_{\mathcal{G}(X_j^{/\mathcal{S}}) = \bold g}\\
   \widehat{ \Pr}_{\textsc{T2I}(\textsc{BERT}(p_{/\mathcal{S}  \cup \{i\}}))}[\mathcal{G}(X) = g] & \triangleq \frac{1}{m} \sum_{j = 1}^{m} 1_{\mathcal{G}(X_j^{/\mathcal{S}  \cup \{i\}}) = \bold g}, 
\end{align}
where $1_{f(x) = \bold g} = 1$ if $f(x) = \bold g $ and $0$ otherwise. 

\subsection*{Proof of Theorem \ref{thm::wordInfluenceConcentrates}}
\begin{proof}
    \begin{align*}
        & \Pr[| \widehat{TI}(p, i, r, \bold g) - {TI}(p, i, r, \bold g)| > t] \\
        & =  \Pr[| \sum_{\substack{\mathcal{S} \subseteq [k] \\ i \not \in \mathcal{S} \\  |\mathcal{S}| \leq r-1}} \frac{\widehat{ \Pr}_{\textsc{T2I}(\textsc{BERT}(p_{/\mathcal{S}}))}[\mathcal{G}(X) = g]- \widehat{ \Pr}_{\textsc{T2I}(\textsc{BERT}(p_{/\mathcal{S}  \cup \{i\}}))}[\mathcal{G}(X) = g]}{{k-1 \choose |\mathcal{S}|}} - {TI}(p, i, r, \bold g)| > t] \\
    \end{align*}
For simplicity, denote $\widehat{ \Pr}_{\textsc{T2I}(\textsc{BERT}(p_{/\mathcal{S}}))}[\mathcal{G}(X) = g]$ by $\widehat{P}_{\mathcal{S}}$ and $ \widehat{ \Pr}_{\textsc{T2I}(\textsc{BERT}(p_{/\mathcal{S}  \cup \{i\}}))}[\mathcal{G}(X) = g]$ by $\widehat{P^*}_{\mathcal{S}}$. Therefore, we can write:

\begin{align*}
        & \Pr[| \widehat{TI}(p, i, r, \bold g) - {TI}(p, i, r, \bold g)| > t] \\
        & =  \Pr[| \sum_{\substack{\mathcal{S} \subseteq [k] \\ i \not \in \mathcal{S} \\  |\mathcal{S}| \leq r-1}} \frac{\widehat{P}_{\mathcal{S}} - \widehat{P^*}_{\mathcal{S}}  - {P}_{\mathcal{S}} + {P^*}_{\mathcal{S}}} {{k-1 \choose |\mathcal{S}|}} > t] \\
        & \leq \sum_{\substack{\mathcal{S} \subseteq [k] \\ i \not \in \mathcal{S} \\  |\mathcal{S}| \leq r-1}} \Pr[| \frac{\widehat{P}_{\mathcal{S}} - \widehat{P^*}_{\mathcal{S}}  - {P}_{\mathcal{S}} + {P^*}_{\mathcal{S}}} {{k-1 \choose |\mathcal{S}|}} > \frac{t}{(\sum_{|\mathcal{S}|}  {k-1 \choose |\mathcal{S}|})}]\\
        & \leq \sum_{\substack{\mathcal{S} \subseteq [k] \\ i \not \in \mathcal{S} \\  |\mathcal{S}| \leq r-1}} \Pr[| \frac{\widehat{P}_{\mathcal{S}} - {P}_{\mathcal{S}} } {{k-1 \choose |\mathcal{S}|}} > \frac{t}{2(\sum_{|\mathcal{S}|}  {k-1 \choose |\mathcal{S}|})}] + \Pr[| \frac{\widehat{P}_{\mathcal{S}}^* - {P}_{\mathcal{S}}^* } {{k-1 \choose |\mathcal{S}|}} > \frac{t}{2(\sum_{|\mathcal{S}|}  {k-1 \choose |\mathcal{S}|})}]\\
        & \leq 4 \sum_{\substack{\mathcal{S} \subseteq [k] \\ i \not \in \mathcal{S} \\  |\mathcal{S}| \leq r-1}}  \exp^{-\frac{t^2 m {k-1 \choose |\mathcal{S}|}^2 }{ (\sum_{|\mathcal{S}|}  {k-1 \choose |\mathcal{S}|})^2 }}
\end{align*}
Where the last inequality comes from Hoeffding's inequality. 
\end{proof}

% \subsection*{Text Transmutations} 

\begin{table}[h]
    \centering
    \begin{tabular}{|c|c|}
       \hline
       Masked Word & Transmutation \\
       \hline
       A & \textit{\underline{the}} respected doctor at the hospital \\
        & \textit{\underline{highly}} respected doctor at the hospital \\
        & \textit{\underline{well}} respected doctor at the hospital \\
       \hline
       respected & a \textit{\underline{staff}} doctor at the hospital \\
	& a \textit{\underline{female}} doctor at the hospital \\
	& a \textit{\underline{retired}} doctor at the hospital \\
       \hline
       doctor & a respected \textit{\underline{physician}} at the hospital \\
	& a respected \textit{\underline{surgeon}} at the hospital \\
	& a respected \textit{\underline{official}} at the hospital \\
       \hline
       hospital & a respected doctor at the \textit{\underline{university}} \\
	& a respected doctor at the \textit{\underline{clinic}} \\
	& a respected doctor at the \textit{\underline{time}} \\
       \hline
    \end{tabular}
    \vspace{1em}
    \caption{Example transmutations for the prompt "A respected doctor at the hospital"}
    \label{table:trasmutations}
\end{table}

\subsection*{CLIP Classifier on CelebA}

CelebA (Celebrities Attributes) \cite{liu2015faceattributes} is a large-scale face attribute dataset containing over 200,000 celebrity images, each with 40 attribute annotations including sex, age, facial expression, and more. The sex attribute takes one of two possible values: "Male" or "Female". Following is the classification report of CLIP classifier on CelebA dataset. The results \ref{clip1} of this experiment suggest that CLIP when used as a sex classifier is fairly accurate.\\

\begin{table}[h]
\centering
\label{tab:clipclassifierceleba}
\begin{tabular}{|c|c|c|c|c|}
\hline
\textbf{Class} & \textbf{Precision} & \textbf{Recall} & \textbf{F1 Score} & \textbf{Support} \\ \hline
female & 0.97 & 1.00 & 0.98 & 94509 \\ \hline
male & 1.00 & 0.96 & 0.98 & 68261 \\ \hline
accuracy & & & 0.98 & 162770 \\ \hline
macro average & 0.98 & 0.98 & 0.98 & 162770 \\ \hline
weighted average & 0.98 & 0.98 & 0.98 & 162770 \\ \hline
\end{tabular}
\vspace{1em}
\caption{Performance of CLIP classifier on sex attribute of CelebA dataset} \label{clip1}
\end{table}

\subsection*{CLIP Classifier on generated images}

We also benchmark CLIP classifier against a pre-trained image classification model DeepFace \cite{serengil2020lightface} with two labels ('male' and 'female'). As we are interested in CLIP's performance on generated images, we run the classifiers on images generated with the prompt "a respected doctor at the hospital" and it's transmutations. Furthermore, a generated image could have multiple people, hence, we first pass the image through an object detector \cite{DBLP:journals/corr/abs-2106-00666} to identify the bounding boxes of people, and run both the classifiers on cropped bounding boxes. The results are shown in Table \ref{clip2}. We can see that CLIP as a classifier is consistent with other image classifiers, suggesting that CLIP is largely unbiased, and the observed bias lies in the image generation process.

\begin{table}[h]
\centering
\label{tab:clipclassifiergenerated}
\begin{tabular}{|c|c|c|c|c|}
\hline
\textbf{Class} & \textbf{Precision} & \textbf{Recall} & \textbf{F1 Score} & \textbf{Support} \\ \hline
female & 0.36 & 0.92 & 0.52 & 13 \\ \hline
male & 0.99 & 0.83 & 0.91 & 127 \\ \hline
accuracy & & & 0.84 & 140 \\ \hline
macro average & 0.68 & 0.88 & 0.71 & 140 \\ \hline
weighted average & 0.93 & 0.84 & 0.87 & 140 \\ \hline
\end{tabular}
\vspace{1em}
\caption{Performance of CLIP as a classifier and DeepFace on generated images with the prompt "a well respected doctor at the hospital". Note that the support for male and female images is vastly different due to the bias in generation process.} \label{clip2}
\end{table}

\subsection*{Values for Placeholder in Large-Scale Evaluation (Section \ref{sec:large-scale})}
$\text{\emph{[ADJECTIVE]}}$ is a placeholder for one of $\{$\emph{``confident''}, \emph{``caring''}, \emph{``rational''}, \emph{``sensitive''}, \emph{``smart''}, \emph{``successful''}$\}$; $\text{\emph{[PERSON]}}$ is a placeholder for one of $\{$\emph{``doctor''}, \emph{``scientist''}, \emph{``artist''}, \emph{``nurse''}, \emph{``lawyer''}$\}$; and $\text{\emph{[PLACE]}}$ is a placeholder for one of $\{$\emph{``office''}, \emph{``gym''}, \emph{``salon''}, \emph{``spa''}, \emph{``mall''}$\}$.
%%%%%%%%%%%%%%%%%%%%%%%%%%%%%%%%%%%%%%%%%%%%%%%%%%%%%%%%%%%%%%%%%%%%%%%%%%%%%%%
%%%%%%%%%%%%%%%%%%%%%%%%%%%%%%%%%%%%%%%%%%%%%%%%%%%%%%%%%%%%%%%%%%%%%%%%%%%%%%%

\end{document}